\documentclass[graybox,a4paper]{svmult}
\usepackage{geometry}
\usepackage{graphicx}
\usepackage{multicol}

\usepackage{color}
\usepackage{enumitem}

\usepackage{amsmath}
\usepackage{amssymb}
\usepackage{siunitx}
\usepackage{newtxtext,newtxmath}  
\usepackage{helvet}         
\usepackage{courier}        
\usepackage{pifont}      

\usepackage{booktabs}
\usepackage{tabularx}
\newcolumntype{Y}{>{\centering\arraybackslash}X}
\newcolumntype{Z}{>{\raggedleft\arraybackslash}X}

\usepackage{adjustbox}

\usepackage{xurl}
\usepackage[hang,flushmargin,symbol]{footmisc}

\definecolor{dark-green}{RGB}{12,80,12}

\usepackage{multirow}

\newcommand{\secref}[1]{Section~\ref{#1}}
\renewcommand{\eqref}[1]{Equation~(\ref{#1})}
\newcommand{\figref}[1]{Figure~\ref{#1}}
\newcommand{\tabref}[1]{Table~\ref{#1}}

\newcolumntype{?}{!{\vrule width 1pt}}

\usepackage{titlesec}
\titlespacing*{\section}{0pt}{1.2\baselineskip}{\baselineskip}
\titlespacing*{\subsection}{0pt}{1.2\baselineskip}{\baselineskip}

\makeindex             

\begin{document}
%
\title*{Learning Long-Horizon Robot Exploration Strategies for Multi-Object Search in\\Continuous Action Spaces}
%
\titlerunning{Learning Long-Horizon Robot Exploration}  
%
 \author{Fabian Schmalstieg$^*$ \and Daniel Honerkamp$^*$ \and Tim Welschehold \and Abhinav Valada}
 \authorrunning{Fabian Schmalstieg \and Daniel Honerkamp \and Tim Welschehold \and Abhinav Valada}
\institute{$^*$These authors contributed equally.\\
All authors are with the Department of Computer Science, University of Freiburg, Germany,\\
This work was funded by the European Union’s Horizon 2020 research and innovation program under grant agreement No 871449-OpenDR.
}

\maketitle              

\abstract{Recent advances in vision-based navigation and exploration have shown impressive capabilities in photorealistic indoor environments. However, these methods still struggle with long-horizon tasks and require large amounts of data to generalize to unseen environments. In this work, we present a novel reinforcement learning approach for multi-object search that combines short-term and long-term reasoning in a single model while avoiding the complexities arising from hierarchical structures. In contrast to existing multi-object search methods that act in granular discrete action spaces, our approach achieves exceptional performance in continuous action spaces. We perform extensive experiments and show that it generalizes to unseen apartment environments with limited data. Furthermore, we demonstrate zero-shot transfer of the learned policies to an office environment in real world experiments.}

\section{Introduction}\label{sec:intro}
Exploration and navigation of unmapped 3D environments is an important task for a wide range of applications across both service and industrial robotics. Research in Embodied AI has made substantial progress in integrating high-dimensional observations in a range of navigation and exploration tasks~\cite{savva2019habitat, chaplot2020learning, chen2018learning}. Recent work has introduced several tasks around multi-object search and exploration \cite{fang2019scene, wani2020multion}. These tasks are particularly challenging in unmapped environments, as they require balancing long-term reasoning about where to go with short-term control and collision avoidance. Without prior knowledge of a floor plan, there is often no obvious optimal policy. 

Furthermore, the combination of complex observation space and long horizons remains an open problem with success rates quickly decreasing as the distance to the goal or the number of objects in the task increases.

Previous work has in particular focused on constructing rich map representations and memories for these agents \cite{fang2019scene, wani2020multion, vodisch2022continual} and demonstrated their benefits while acting in granular discrete action spaces. The long-horizon nature of these tasks poses a significant challenge for learning-based methods, as they struggle to learn longer-term reasoning and to explore efficiently over long horizons. 
This problem is strongly exacerbated in fine-grained continuous action spaces. A common strategy to mitigate this challenge is to instead learn high-level waypoints which are then fed to a low-level controller \cite{chaplot2020object, chen2020learning}. While this can simplify the learning problem by acting in a lifted MDP with a much shorter horizon length, it limits the ability of the agent to simultaneously learn to control other aspects such as its camera or arms at a higher control frequency. During inference, high-level actions can be taken at arbitrary frequencies by executing them in a model predictive control style manner but the agent is bound to a low control frequency during training.

\begin{figure}[t]
	\centering
	\resizebox{\columnwidth}{!}{%
  		\includegraphics[bb=0 0 1280 960, width=0.32\columnwidth,trim={0.0cm 0.0cm 0.0cm 0.0cm},clip,angle =0]{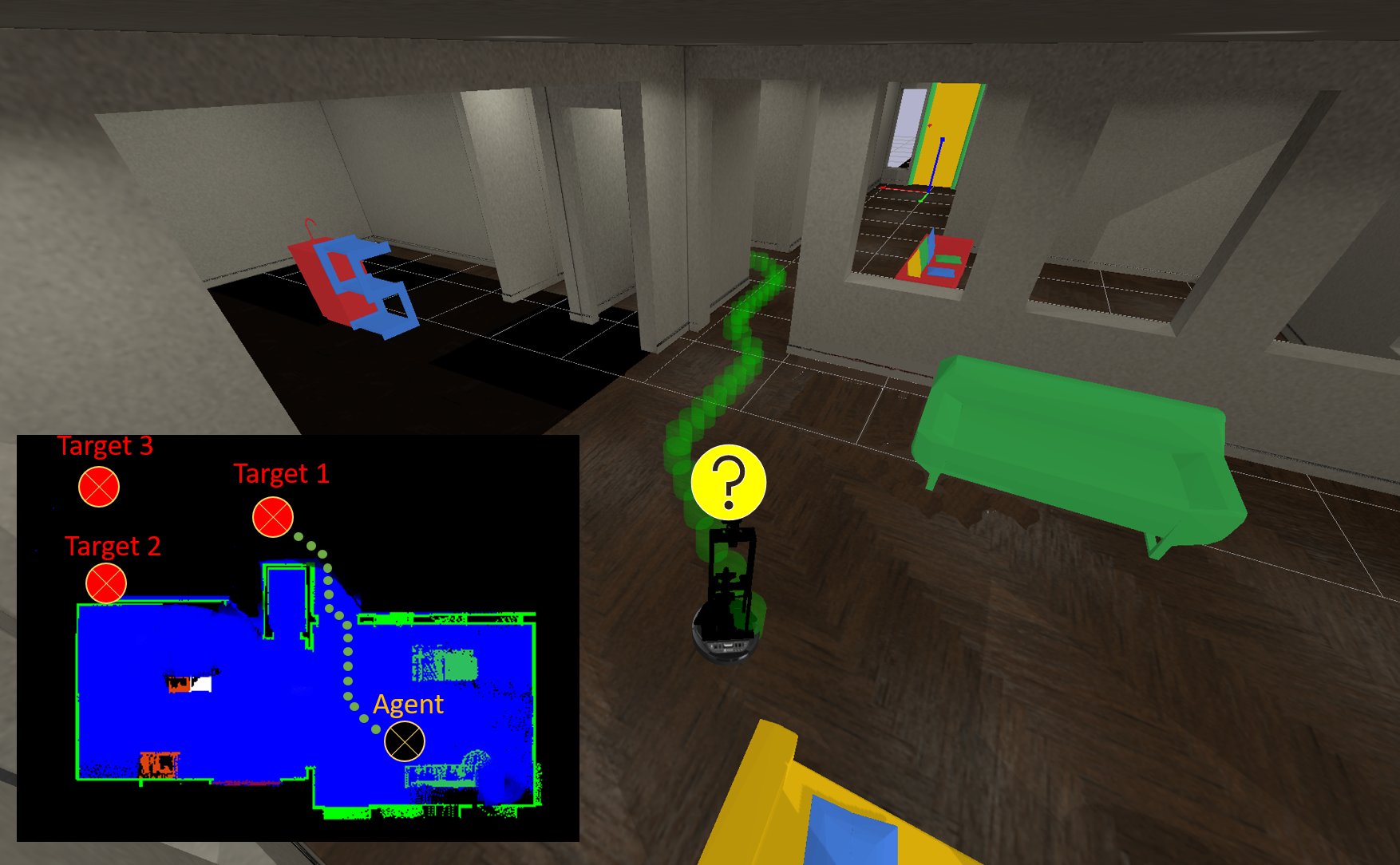}
	}
	\caption{Starting in an unexplored map and given a set of target objects, the robot faces the complex decision on how to most efficiently find these objects. Our approach continuously builds a semantic map of the environment and learns to combine long-term reasoning with short-term decision making into a single policy by predicting the direction of the path towards the closest target object, shown in green. Note that the agent does not receive the path consisting of the waypoints or the location of the objects.}

  	\label{fig:intro}
\end{figure}
In this work, we present a novel approach to reason about long-horizon exploration while learning to directly act in continuous action spaces at high control frequencies and demonstrate its effectiveness in multi-object search tasks.

Our approach learns to predict the direction of the path towards the closest target object. It then learns a policy that observes this prediction, enabling it to express long-term intentions while taking short-term actions based on the full context. As a consequence, the policy can incorporate expected inaccuracies and uncertainties in the predictions and balance strictly following its intentions with short-term exploration and collision avoidance. \figref{fig:intro} illustrates the task and our approach.

To jointly learn this behavior with a single model, we introduce a learning curriculum that balances groundtruth and learned intentions. In contrast to many learning-based navigation approaches building on complex features from color or depth images that make generalization to unseen environments hard and data intensive~\cite{hurtado2021learning}, we learn purely from structured, semantic map representations that have more universal features and are therefore more independent of the specific training environment and show only minimal sim-to-real gap. We perform extensive experimental evaluations and demonstrate that our proposed approach achieves strong generalization to unseen environments from a very small number of training scenes and directly transfers to the real world. Lastly, the combination of these simple inputs and expressive navigation intentions ensures good interpretability of the agent's decisions and failures. 
To the best of our knowledge, this is the first real-world demonstration of learning-based multi-object search tasks.

To summarise, this paper makes the following contributions:
\begin{itemize}[topsep=0pt]
    \item We present a novel search agent that unifies long-horizon decision making and frequent low-level control into a single time-scale and model.
    \item We propose the first multi-object search task in continuous action space.
    \item We demonstrate that our approach is capable to learn from very limited training data and achieves strong generalization performance in unseen apartment environments and zero-shot transfer to the real world.
    \item We make the code publicly available at \url{http://multi-object-search.cs.uni-freiburg.de}.
\end{itemize}

\section{Related Work}\label{sec:related}

{\parskip=5pt\noindent\textit{Embodied AI tasks}:}
Navigation in unmapped 3D environments has attracted a lot of attention in recent work on embodied AI, covering a range of tasks. In PointGoal navigation~\cite{savva2019habitat, chaplot2020learning}, the agent at each step receives the displacement vector to the goal that it has to reach. Whereas, in AudioGoal navigation~\cite{chen2020soundspaces, younes2021catch}, the agent at each step receives an audio signal emitted by a target object. Conversely, in ObjectGoal navigation~\cite{zhu2017target, chaplot2020object, qiu2020learning, druon2020visual}, the agent receives an object category that it has to navigate to.

Extending the ObjectGoal task, Beeching~\textit{et~al.}~\cite{beeching2021deep} propose an ordered $k$-item task in a Viz Doom environment, in which the agent has to find $k$ items in a fixed order. Similarly in the MultiOn task~\cite{wani2020multion} the agent has to find $k$ objects in realistic 3D environments. Fang~\textit{et~al.}~\cite{fang2019scene} propose an object search task in which the agent has to find $k$ items in arbitrary order. Here the target object locations are defined by the given dataset, i.e. are in a fixed location in each apartment. In contrast, we use a random distribution over the target locations, strongly increasing the diversity. While all of these works focus on discrete actions (move forward, turn left, turn right, stop), we show that our approach can directly learn in the much larger continuous action space and demonstrate the direct transfer to the real world.

{\parskip=5pt\noindent\textit{Object search and exploration}:}
Approaches for multi-object search and navigation tasks fall into two categories:
implicit memory agents and agents that explicitly construct a map of the environment as a memory.
Agents without an explicit map include direct visual end-to-end learning from RGB-D images as well as FRMQN~\cite{oh2016control} and SMT~\cite{fang2019scene} which store an embedding of each observation in memory, then retrieve relevant information with an attention mechanism.
On the other side of the spectrum, we have agents that project the RGB-D inputs into a global map, building up an explicit memory. They then commonly extend this representation with semantic object annotations~\cite{henriques2018mapnet, beeching2021deep}. Wani~\textit{et~al.}~\cite{wani2020multion} provide a comprehensive comparison of these methods. While they train separate agents for each number of target objects, we train a single agent that generalizes to different numbers of target objects.

SGoLAM~\cite{kim2021sgolam} combine a mapping and a goal detection module. Then either employ frontier exploration if no goal object is in sight or if a goal is detected, they use a D*-planner to move closer to the goal. They achieve strong results without any learning component. Closely related to object search, previous work has also focused on pure exploration of realistic apartments. This includes frontier based approaches~\cite{yamauchi1997frontier} as well as reinforcement learning with the aim to maximise coverage~\cite{chen2018learning}.

{\parskip=5pt\noindent\textit{Learning long horizon goals}:}
Multi-object search and exploration with an embodied agent combine long-horizon thinking with short-horizon control to navigate and avoid collisions. This can pose a challenge for learning-based approaches. This has previously been mitigated by learning higher-level actions at a lower control frequency such as learning to set waypoints or to directly predict task goals \cite{min2022film}, which then get passed down to a lower-level planner for navigation~\cite{chaplot2020object, chen2020learning}. While this shortens the horizon of the MDP the agent is acting in, it makes it difficult to learn to simultaneously control other aspects at a lower time scale, such as controlling a camera joint or a manipulator arm. Our approach directly learns at a high control frequency and as such can directly be extended to integrate such aspects. In our experiments, we furthermore demonstrate that direct prediction of the goal locations does not generalize well for multi-object search. 
The long horizon problem is further exacerbated in continuous control tasks. While most existing work focuses on granular discrete actions~\cite{fang2019scene, wani2020multion, chaplot2020object, kim2021sgolam}, our approach succeeds in a continuous action space.

{\parskip=5pt\noindent\textit{Mapping}:}
Spatial maps built with Simultaneous Localization and Mapping (SLAM) have been used for tasks such as exploration~\cite{zhang2017neural,cattaneo2022lcdnet} and FPS games~\cite{beeching2021deep}. Both occupancy and semantic maps have commonly been used in embodied AI tasks~\cite{wani2020multion, kim2021sgolam} and several works have presented approaches to build such maps in complex environments~\cite{chaplot2020learning, chaplot2020object}. In our approach, we assume access to a method to build such maps.

\section{Learning Long-Horizon Exploration}\label{sec:approach}
In this section, we first define the multi-object search task and formulate it as a reinforcement-learning problem in \secref{sec:problem}. We then introduce our approach for learning a novel predictive task and an effective method to jointly learn low- and high-level reasoning in \secref{sec:agent}.

\subsection{Problem Statement}
\label{sec:problem}

In each episode, the agent receives a list of up to $k$ object categories, drawn randomly from a total set of $c$ categories. It then has to search and navigate to these objects which are spawned randomly within an unmapped environment. An object is considered found if the agent has seen it and navigates up to a vicinity of \SI{1.3}{\meter} of the target object. 
In contrast to MultiOn~\cite{wani2020multion}, we require a single agent to learn to find variable numbers of target objects and focus on unordered search, meaning the agent can find these objects in any order it likes. On one hand, this provides the agent with more freedom. On the other hand, the optimal shortest-path policy is non-trivial, even in a mapped environment, making this a very challenging task to solve optimally.

This can be formulated as a goal-conditional Partially Observable Markov Decision Process (POMDP) $\mathcal{M} = (\mathcal{S}, \mathcal{A}, \mathcal{O}, T(s' | s, a), P(o | s), R(s, a, g))$, where $\mathcal{S}$, $\mathcal{A}$ and $\mathcal{O}$ are the state, action and observation spaces, $T(s' |s, a)$ and $P(o | s)$ describe the transition and observation probabilities and $R(s, a, g)$ is the reward function.
At each step, the agent receives a visual observation $o$ from an RGB-D and semantic camera together with a binary vector $g$ indicating which objects it must find. Its aim is to learn a policy $\pi(a | o, g)$ that maximises the discounted, expected return $\mathbb{E}_\pi[\sum_{t=1}^{T} \gamma^t R(s_t, a_t, g)]$, where $\gamma$ is the discount factor. The agent acts in the continuous action space of a LoCoBot robot, controlling the linear and angular velocities of its differential drive. During training, these actions are executed at a control frequency of \SI{10}{\hertz}. The agent receives a sparse reward of 10 whenever it finds a target object. It furthermore receives a dense time penalty of $-0.0025$ per step, a distance reward for getting closer to the next target object, and a collision penalty of -0.1. An episode ends after successfully finding all objects, exceeding 600 collisions or exceeding 3500 steps.

\subsection{Technical Approach}
\label{sec:agent}

\begin{figure}
	\centering
	\resizebox{\columnwidth}{!}{%
  		\includegraphics[width=0.32\columnwidth,trim={0.0cm 0.0cm 0.0cm 0.0cm},clip,angle =0]{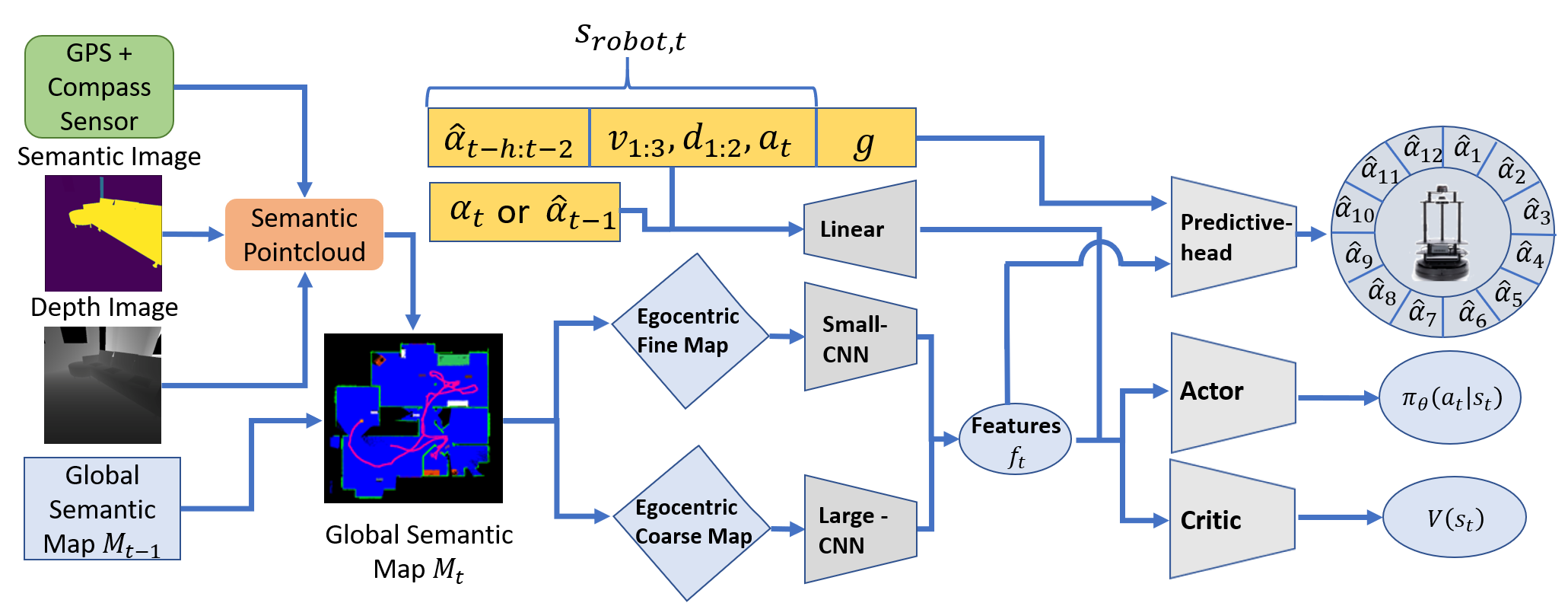}
	}
	\caption{Our proposed model architecture. The mapping module aggregates depth and semantic information into a global map. The predictive module learns long-horizon relationships which are then interpreted by a reinforcement learning policy. During training, the agent receives a state vector with either the groundtruth direction to the closest object $\alpha_t$ or its prediction $\hat{\alpha_t}$. At test time it always receives its prediction. It furthermore receives its previous predictions $\hat{\alpha}_{t-18:t-2}$, the variances of its x- and y-position $v_{1}$ and $v_{2}$, the circular variance of its predictions $v_3$, a collision flag $d_1$, the sum over the last 16 collisions $d_2$, its previous action $a_t$, and a binary vector $g$ indicating the objects the agents have to find. 
	}
  	\label{fig:architecture}
\end{figure}

We propose a reinforcement learning approach that consists of three components: a mapping module, a predictive module to learn long-horizon intentions, and a reinforcement learning policy. The full approach is depicted in \figref{fig:architecture}.

{\parskip=5pt\noindent\textit{Mapping}:}
The mapping module uses the $128\times128$ pixels depth and semantic camera to project the points into a local top-down map. From this, it then updates an internal global map which is further annotated with the agent's trace and encoded into a standard RGB image. In each step, the agent then extracts an egocentric map from it and passes two representations of this map to the encoder: a coarse map of dimension $224 \times 224 \times 3$ at a resolution of $\SI{6.6}{\centi\meter}$ and a fine-grained map of dimension $84 \times 84 \times 3$ at a resolution of $\SI{3.3}{\centi\meter}$. Meaning they cover $\SI{14.8}{\meter} \times \SI{14.8}{\meter}$ and $\SI{2.77}{\meter} \times \SI{2.77}{\meter}$, respectively. After the agent has segmented an object correctly, it updates the object's annotation to a fixed color-coding to mark the corresponding object as “found”.
The coarse map is then encoded into a 256-dimensional feature vector and the fine map into a 128-dimensional feature vector using a convolutional neural network, before being concatenated into the feature embedding $f_t$. The coarse map is passed through a ResNet-18~\cite{he2016deep} pre-trained on ImageNet~\cite{deng2009imagenet}, while the local map is encoded by a much simpler three-layer CNN with 32, 64, and 64 channels and strides 4, 2, and 1. 

{\parskip=5pt\noindent\textit{Learning Long-horizon Reasoning}:}
While directly learning low-level actions has shown success in granular discrete environments~\cite{fang2019scene, wani2020multion}, this does not scale to continuous environments as we show in \secref{sec:experiments}. We hypothesize that this is due to being unable to explore the vast state-action space efficiently.
To learn long-horizon reasoning within a single model, we introduce a predictive task to express the agent's navigation intentions.
In particular, the agent learns to predict the direction of the shortest path to the currently closest target object. It does so by estimating the angle to a waypoint generated by an $A^*$ planner in a distance of roughly $\SI{0.4}{\meter} - \SI{0.55}{\meter}$ from the agent (varying due to the discretized grid of the planner) as illustrated in \figref{fig:intro}. The plan is generated in a map inflated by \SI{0.2}{\meter} to avoid waypoints close to walls or obstacles.

The prediction is parameterized as a neural network $\vec{\hat{\alpha}} = f_{pred}(f_t, s_{robot, t}, g)$ that takes a shared feature encoding $f_t$ from the map encoding, the robot state $s_{robot, t}$ without the groundtruth vector $\alpha_t$ nor the prediction  $\hat{\alpha}_t$ and the goal objects $g$ and predicts a vector of probabilities over the discretized angles to this waypoint. In particular, we discretize the angle into 12 bins and normalize the outputs with a softmax function.

This task is used both in the form of an auxiliary task to shape representation learning, propagating gradients into a shared map encoder, as well as a recursive input to the agent, allowing it to guide itself as the agent observes the previous step's predictions both as an overlay in the ego-centric map and the robot state.

The predictive module minimizes the cross-entropy between the predictions and a one-hot encoding of the groundtruth angle $\vec{\alpha}$ given by
\begin{equation}\label{eq:auxloss}
    \mathcal{L}_{pred} = \frac{1}{N} \sum_{i=1}^{N} \vec{\alpha}_i \log \hat{\vec{\alpha}}_i.
\end{equation}

The discrete distribution enables the agent to cover multi-modal hypotheses, for example when standing on a floor with several unexplored directions. As a result, the prediction can vary more smoothly over time in contrast to commonly used uni-modal distributions such as a Gaussian, which can fluctuate rapidly if the most likely direction changes to another door.

{\parskip=5pt\noindent\textit{Policy}:}
The policy receives the concatenated features $f_t$ of the flattened map encodings, the robot state, and a history of its predictions. The policy is then learned with Proximal Policy Optimization (PPO) \cite{schulman2017proximal}. The actor and critic are parameterized by a two-layer MLP network with 64 hidden units each and a gaussian policy. 
The total loss $\mathcal{L}$ given by
\begin{equation}
    \mathcal{L} = \mathcal{L}_{ppo} + \lambda \mathcal{L}_{pred},
\end{equation}
is jointly minimized with the Adam optimizer \cite{kingma2014adam}.

{\parskip=5pt\noindent\textit{Training}:}
During training we have the choice to provide the policy either with the groundtruth direction or its prediction:
Providing it with the groundtruth vector results in strong coupling between the agent's navigation and the maximum value of the vector, which points to the waypoint corresponding to the closest next object. Hence, when deploying the agent with the auxiliary prediction, the agent blindly follows its predictions. Instead, we want to learn a policy that can, on one hand, react to suboptimal intentions and on the other hand can learn more optimal paths than simply finding the closest object. For such behavior, it is desirable to train the policy directly with its predictions. 

To avoid instabilities from initially very suboptimal predictions, we introduce a learning curriculum that balances observing the groundtruth and the agent's predictions.

The curriculum starts with a probability of 16\% to observe an entire episode with the prediction and otherwise receives the groundtruth. Once the agent has exceeded a success rate of 50\%, we linearly increase this probability by 2\% every 4 episodes up to a maximum of 72\%. This curriculum enables the robot to react to prediction errors and correct its navigation accordingly.
To avoid learning a simple mapping from groundtruth to prediction during training, this $\vec{\alpha}$ is never observed by the predictive module and only passed to the policy.

In specific situations where it is hard to assess a global strategy, the agent sometimes predicts alternating auxiliary angles. We hypothesize that the reason for these highly alternating predictions is that the training episodes run purely with predictions, suffer from suboptimal predictions and navigation and will therefore accumulate more errors contributing to the overall loss.
We test this by disabling the gradient flow from the predictive head back into the rest of the network for the episodes with predictions. We observe that this leads to a lower prediction loss. 
But at the same time the predictions can no longer shape the policy's representations 
and thus, the agent at times struggles with “imperfect” episodes, when deployed. This is reflected, in a tremendous drop in the success rate when executing prediction episodes. We conclude that the gradient flow from all episodes is a crucial component.

With the focus on structured map inputs and the learning curriculum enables, we are able to train the agent in the comparatively small number of 4,500,000 steps.

\section{Experimental Evaluations}\label{sec:experiments}
To demonstrate the effectiveness of our approach, we perform extensive evaluations in simulation and the real world.
With these experiments we aim to answer the following questions:
\begin{itemize}
    \item Does the learned behavior generalize to unseen environments?
    \item Does the agent learn to efficiently use the long-term prediction? Does this lead to more efficient exploration than using alternative approaches?
    \item Does the learned behavior generalize to the real world?
\end{itemize}

\subsection{Experimental Setup}
We train a LoCoBot robot in the iGibson environment. The LoCoBot has a differential drive and is equipped with an RGB-D camera with a field of view of 79 degrees and a maximum depth of \SI{5.6}{\meter}. The action space consists of the linear and angular velocities for the base. We construct tasks of finding 1-6 target objects, matching the hardest setting in previous work \cite{wani2020multion}. We use the same eight training scenes as the iGibson challenge\footnote{\url{http://svl.stanford.edu/igibson/challenge2020.html}} and use the remaining seven unseen apartments for evaluation.  

As larger datasets such as Matterport3D or Gibson currently do not support semantic camera observations, we leave evaluations on these to future work. The PPO agent is based on an open-source implementation \cite{stable-baselines3}. 

{\parskip=5pt\noindent\textit{Evaluation Metrics}:}
We focus on two metrics: the ability to find all target objects, defined by the success rate, and the optimality of the search paths, measured by the success weighted by Path Length (SPL) \cite{anderson2018evaluation}.
We evaluate each scene for 75 episodes, which results in a total of 600 episodes for the train and 525 for the test set for each approach. For the learning based approaches we evaluate the best performing checkpoint on the training scenes. We report the mean over two training seeds for our approach.

\setlength{\tabcolsep}{6pt}
\begin{table}[t]
  \centering
  \caption{Hyperparameters used for training. One sensitive parameter is ppo epoch in combination with the clip parameter. Setting the parameter too high causes some behaviour which is similar to catastrophic forgetting.}
  \label{tab:hyper}
  
  \begin{tabular}{l|c?l|c}
  \toprule
Parameter & Value & Parameter & Value \\
\midrule
clip param      & 0.1      & $\gamma$        & 0.99\\
ppo epoch       & 4        & lr              & 0.0001\\
num mini batch  & 64       & max grad norm   & 0.5\\
value loss coef & 0.5      & optimizer       & Adam \\
entropy coef    & 0.005    && \\

\bottomrule

  \end{tabular}
  
\end{table}

{\parskip=5pt\noindent\textit{Baselines}:}
To test the effectiveness of learning long-range navigation intentions, we compare our approach against a range of action parametrizations as well as a recent non-learning based state-of-the-art approach.\\
{\parskip=5pt\noindent\textit{Map-only}} represents the standard end-to-end reinforcement learning approach and conceptually matches competitive baselines from previous work.  
It receives the same map and robot state inputs as our agent and acts directly in the action space of the differential drive, but does not learn to predict long-horizon intentions. \\

{\parskip=5pt\noindent\textit{Goal-prediction}} Instead of predicting the direction to the next waypoint towards the target, we also evaluate directly predicting the location of the next target object. The agent learns to predict the angle and distance of the next closest target object relative to its current base frame. \\
{\parskip=5pt\noindent\textit{SGoLAM} \cite{kim2021sgolam}} combines non-learning based approaches to achieve very strong performance on the CVPR 2021 MultiOn challenge. It explores the map with frontier exploration until it localizes a target object, then switches to a planner to navigate to the target. We reimplement the author's approach for continuous action spaces, closely following the original implementation where possible. While the original implementation relies on two threshold values, namely $\epsilon$ and $\delta$, for goal localization, we directly use the semantic camera which finds objects more reliably. Due to this, our implementation improves the performance of SGoLAM.

\subsection{Simulation Experiments}
To test the ability to learn long-horizon exploration, we first evaluate the approaches on the seen apartments.  \tabref{tab:seen_combined} reports the results for all approaches across different numbers of target objects. While the map-only approach that purely learns raw continuous actions finds around three quarters of the single objects, the success rates quickly deteriorate with more target objects. We hypothesize that this is due to the agent getting lost in the vast continuous action space, being unable to meaningfully explore the space.
Directly predicting the goal locations slightly improves the performance, but still fails in the majority of cases with five or more objects. This indicates that the agent is not able to meaningfully predict goal locations, as there are a large number of valid hypotheses while the environment is still largely unexplored.

SGoLAM achieves a better performance, yielding an overall success rate of 65.3\%. Note that this approach does not rely on a learning component, therefore it has not encountered any of these apartments before.
In contrast to the other learning-based approaches, our agent is able to consistently solve this task even for six target objects. 
In terms of path optimality measured by the SPL in \tabref{tab:seen_combined}, we observe a similar case across the approaches. While outperforming the other approaches, both ours and SGoLAM achieve low absolute values. However, note that a perfect SPL would require to directly follow the shortest path to all objects and as such is not achievable without knowledge of the object positions. Additionally, one needs to bare in mind that the SPL metric is not guaranteed to find the most efficient path as it computes the path in a greedy manner.

\begin{table}[t]
    \centering
    \caption{Evaluation of \textbf{seen} environments, reporting the success rate (top) and SPL (bottom).}
    \label{tab:seen_combined}
    \begin{tabularx}{\textwidth}{cl|YYYYYY|Y}
      \toprule
        & Model & 1-obj & 2-obj & 3-obj &4-obj & 5-obj & 6-obj & Avg 1-6\\
      \midrule
        \parbox[t]{3mm}{\multirow{4}{*}{\rotatebox[origin=c]{90}{Success}}} 
        & Map-only & 75.0 & 70.6 & 54.7 & 53.3 & 40.3 & 37.2 & 55.1\\
        & Goal prediction & 78.1 & 72.3 & 58.1 & 54.4 & 46.7 & 43.0 & 58.7\\
        & SGoLAM & 82.0 & 77.9 & 62.8 & 60.5 & 58.0 & 51.0 & 65.3\\ 
        & Ours & 95.4 & 90.7 & 89.0 & 87.6 & 85.1 & 83.4 & \textbf{88.5}  \\
        \midrule
        \parbox[t]{3mm}{\multirow{4}{*}{\rotatebox[origin=c]{90}{SPL}}}
        & Map-only & 33.5 & 31.0 & 26.8 & 24.9 & 20.8 & 21.3 & 26.3\\
        & Goal prediction & 31.4 & 29.8 & 25.9 & 23.1 & 18.3 & 22.3 & 25.1 \\
        & SGoLAM & 41.6 & 34.5 & 32.0 & 33.7 & 36.6 & 38.1 & 36.0\\ 
        & Ours  & 46.4 & 40.9 & 42.9 & 44.2 & 49.2 & 53.1 & \textbf{46.1}\\
      \bottomrule
    \end{tabularx}
\end{table}

Subsequently, we evaluate all the agents in unseen environments and present the results in \tabref{tab:unseen_combined}. Interestingly, all learning-based approaches achieve similar performance as on the seen apartments, indicating that the semantic map modality generalizes well to unseen apartments. This is particularly impressive as it only has access to eight training scenes. On the other hand, SGoLAM performs better than in the train split, indicating that the test split might be less challenging. While the gap between the best baseline, SGoLAM, and our approach shrinks, it remains considerable with an average difference in success rates of 6.5\%. In terms of SPL it even performs slightly better. This may be due to the path planner, which, once an object is in sight, executes the optimal path to this object. While SGoLAM achieves very strong performance in apartments with a lot of open areas, its performance drops severely in more complex apartments with many corridors, rooms next to each other, and generally long-drawn layouts. In contrast, our approach maintains a more even performance across the different apartment layouts. 

\begin{table}[t]
    \centering
    \caption{Evaluation of \textbf{unseen} environments, reporting the success rate (top) and SPL (bottom).}
    \label{tab:unseen_combined}
    \begin{tabularx}{\textwidth}{cl|YYYYYY|Y}
      \toprule
        & Model & 1-obj & 2-obj & 3-obj &4-obj & 5-obj & 6-obj & Avg 1-6\\
      \midrule
        \parbox[t]{3mm}{\multirow{4}{*}{\rotatebox[origin=c]{90}{Success}}} 
        & Map-only & 74.2 & 64.5 & 61.0 & 60.7 & 57.9 & 32.5 & 58.4 \\
        & Goal prediction & 74.7 & 69.8 & 66.7 & 61.9 & 56.1 & 44.0 & 62.2\\
        & SGoLAM & 89.8 & 85.3 & 79.9 & 75.0 & 74.3 & 71.1 & 79.2\\
        & Ours & 93.1 & 89.4 & 86.4 & 82.1 & 82.5 & 81.1 &  \textbf{85.7}\\
        \midrule
        \parbox[t]{3mm}{\multirow{4}{*}{\rotatebox[origin=c]{90}{SPL}}} 
        & Map-only & 30.1 & 27.0 & 28.6 & 29.8 & 29.4 & 18.4 & 27.2\\
        & Goal prediction & 29.9 & 21.9 & 20.1 & 21.0 & 19.7 & 21.4 & 22.3\\
        & SGoLAM  & 47.7 & 44.0& 43.7& 46.2 & 47.5 & 49.8 & \textbf{46.4}\\
        & Ours & 45.3 & 40.0 & 38.5 & 43.2 & 46.8 & 49.6 & 43.9\\
      \bottomrule
    \end{tabularx}
\end{table}

\begin{figure}[t]
	\centering
	{\setlength{\fboxsep}{0pt}%
  \setlength{\fboxrule}{0pt}%
	\resizebox{\textwidth}{!}{%
  	\begin{tabular}{ccc}
  		\fbox{\includegraphics[width=0.205\columnwidth,trim={0cm 0cm 0cm 0cm},clip,angle =0]{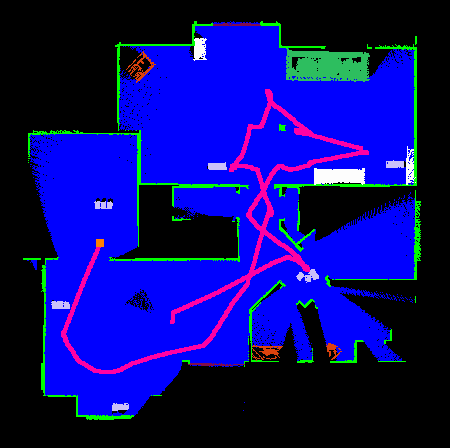}} &
  		\fbox{\includegraphics[width=0.285\columnwidth,trim={0cm 0cm 0cm 0cm},clip,angle =0]{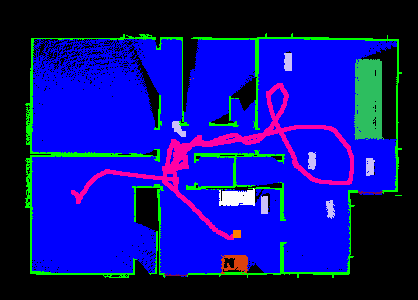}} &
  		\fbox{\includegraphics[width=0.32\columnwidth,trim={0cm 0cm 0cm 0cm},clip,angle =0]{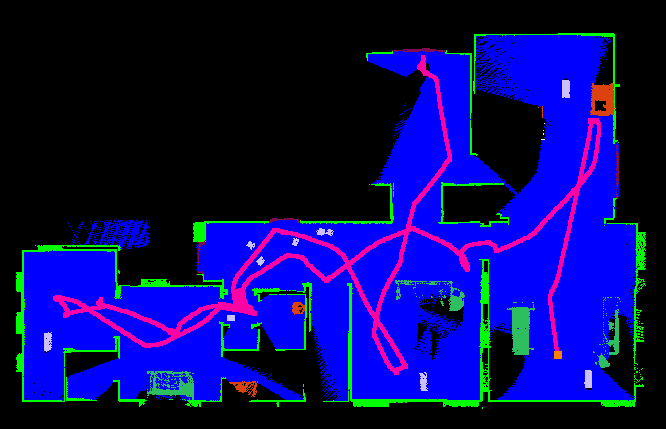}}\\  
  		  		\fbox{\includegraphics[width=0.205\columnwidth,trim={0cm 0cm 0cm 0cm},clip,angle =0]{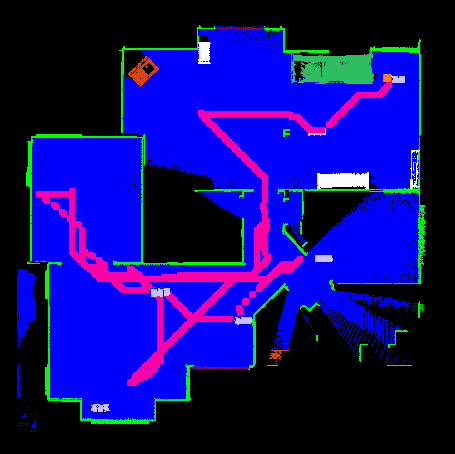}} &
  		\fbox{\includegraphics[width=0.28\columnwidth,trim={0cm 0cm 0cm 0cm},clip,angle =0]{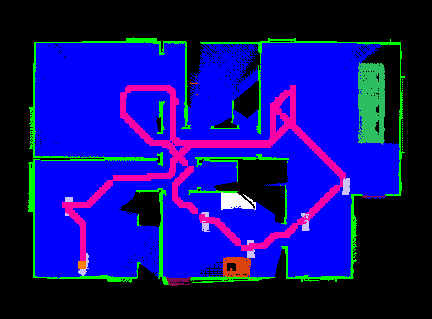}} &
  		\fbox{\includegraphics[width=0.32\columnwidth,trim={0cm 0cm 0cm 0cm},clip,angle =0]{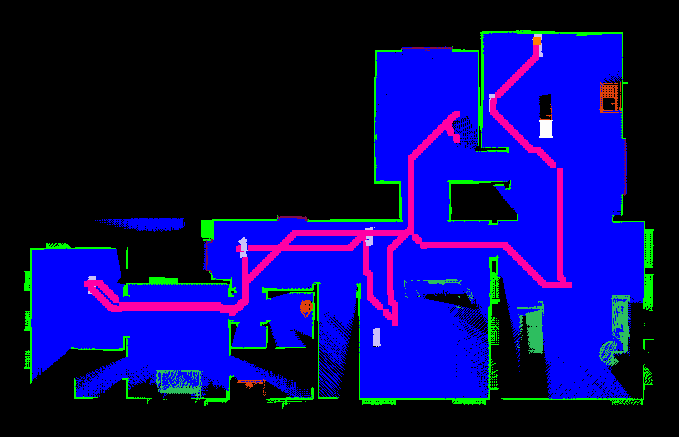}}\\  
	\end{tabular}
	}}
	\vspace{-0.2cm}
	\caption{Example trajectories of our agent (top) and SGoLAM (bottom) in unseen apartments. The maps show the following categories: black: unexplored, blue: free space, green: walls, red: agent trace, grey: (found) target objects, other colors: miscellaneous objects.}
  	\label{fig:example_trajectories}
\end{figure}

Qualitatively, we find that the agent learns efficient navigation behaviors. \figref{fig:example_trajectories} and the accompanying video show example trajectories in unseen apartments. The agent efficiently looks around rooms and learns maneuvers such as a three-point turn. Where confident, it reliably follows its own long-horizon predictions, while deviating if it points into walls, if the predictions are low confidence or if it is possible to explore a lot of additional space with little effort. While SGoLAM randomly picks points on the frontier, often resulting in multiple map crossings and getting lost in very small unexplored spaces, our approach continuously explores and learns to leave out small spots that are unlikely to contain a target object. We further observe an inverse development between the SPL and the success rate with regard to the number of objects for both our approach and SGoLAM. This increase in the SPL most likely stems from a higher number of routes that can be taken which are close to the optimal path. With fewer objects in the scene, large parts of the exploration increase the SPL without finding an object. Nevertheless, this exploration is essential as there is no prior knowledge of the object locations.

\begin{table}[t]
    \centering
    \caption{Real world multi-object search experiments on the HSR Robot.}
    \label{tab:real_world}
    \begin{tabularx}{\textwidth}{l|YYYY|Z}
      \toprule
        Model & 2-obj & 3-obj &4-obj & 6-obj & Total \\
      \midrule
        
        \textbf{Success} & \textbf{~6}      &  \textbf{~6}     & \textbf{~5}     &   \textbf{~5} & \textbf{22} \\
        Collision        & ~4               &  ~3              & ~5              &   ~4          & 16 \\
        Timeout          & ~0               &  ~1              & ~0              &   ~1          & ~2 \\ 
        Total Episodes   & 10               &  10              & 10              &   10          & 40 \\
      \bottomrule
    \end{tabularx}
\end{table}

\setlength{\tabcolsep}{1pt}
\renewcommand{\arraystretch}{1}
\setlength{\tabcolsep}{6pt}
\renewcommand{\arraystretch}{1}

\setlength{\tabcolsep}{1pt}
\renewcommand{\arraystretch}{1}
\begin{figure}
	\centering
	{\setlength{\fboxsep}{0pt}%
  \setlength{\fboxrule}{0pt}%
	\resizebox{\textwidth}{!}{%
  	\begin{tabular}{cc}
  		\fbox{\includegraphics[width=0.5\columnwidth,trim={0cm 0cm 0cm 0cm},clip,angle =0]{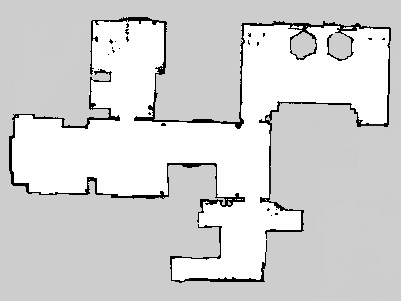}} &
  		\fbox{\includegraphics[width=0.5\columnwidth,trim={0cm 0cm 0cm 0cm},clip,angle =0]{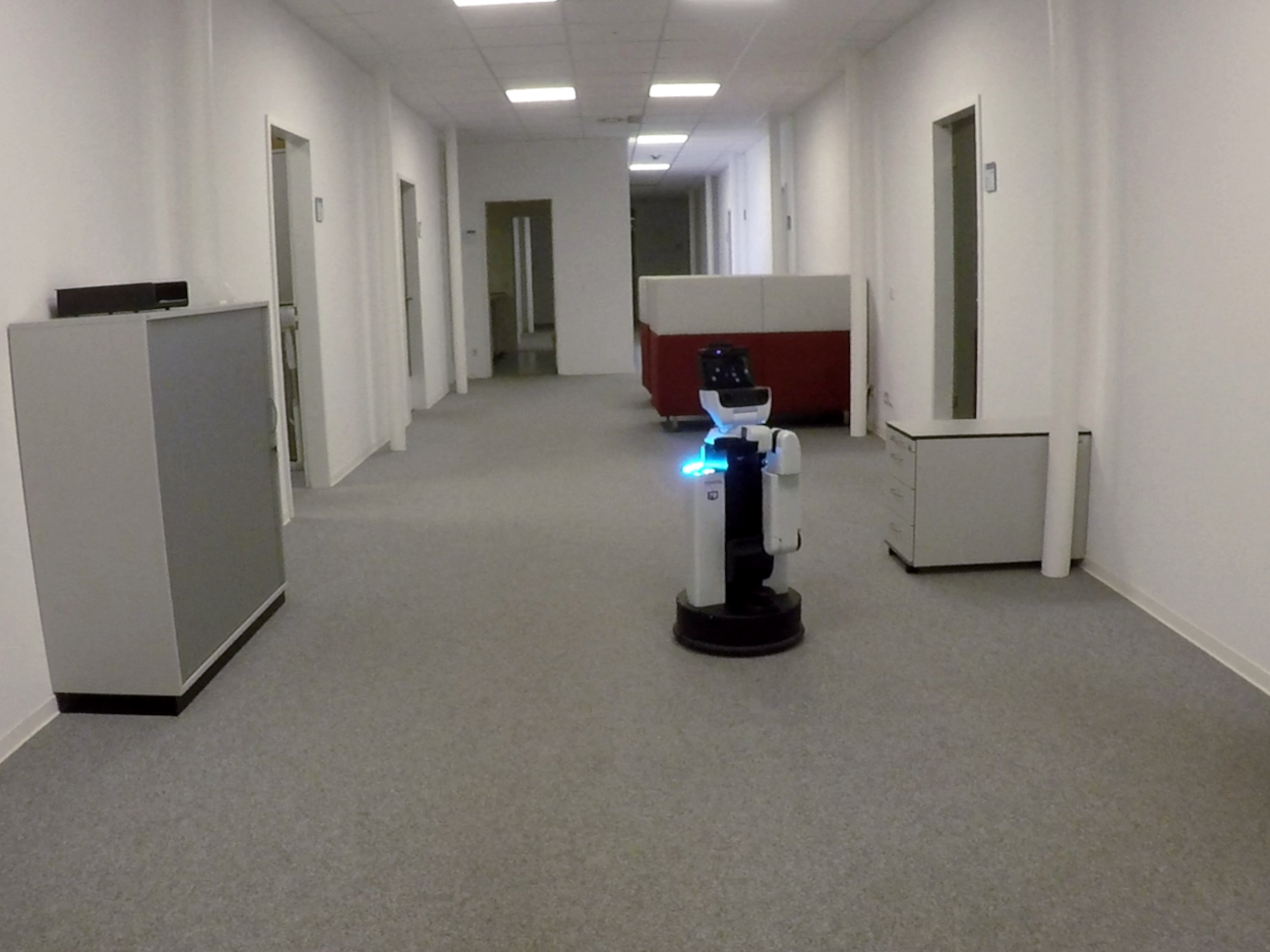}} \\
  		\fbox{\includegraphics[width=0.5\columnwidth,trim={0cm 0cm 0cm 0cm},clip,angle =0]{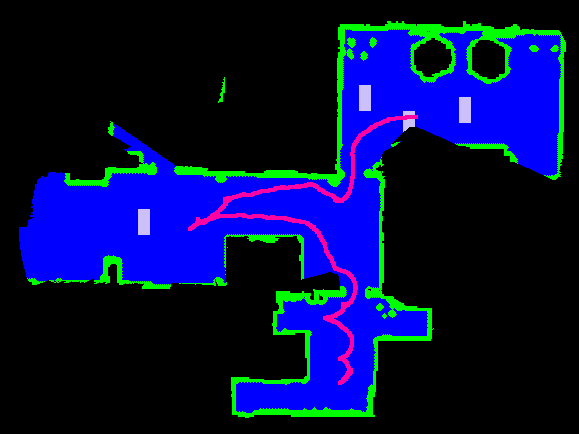}} & 
  		\fbox{\includegraphics[width=0.5\columnwidth,trim={0cm 0cm 0cm 0cm},clip,angle =0]{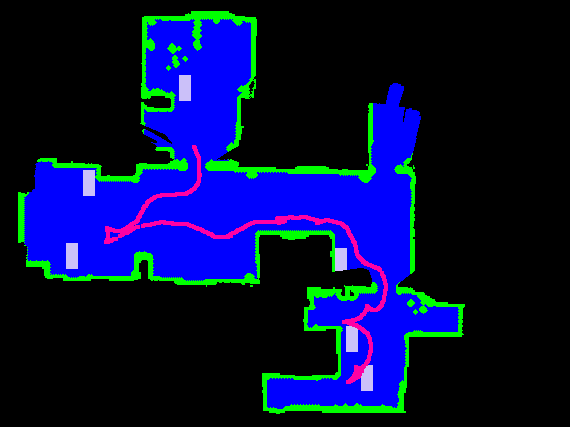}}\\
  		\fbox{\includegraphics[width=0.5\columnwidth,trim={0cm 0cm 0cm 0cm},clip,angle =0]{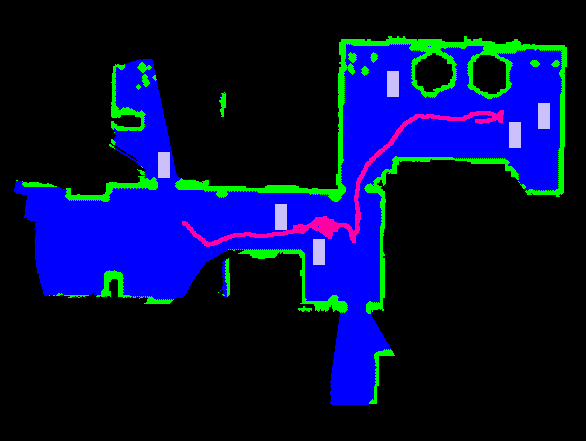}} & 
  		\fbox{\includegraphics[width=0.5\columnwidth,trim={0cm 0cm 0cm 0cm},clip,angle =0]{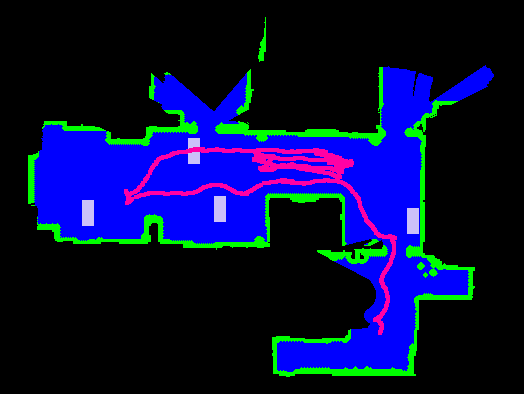}}\\
	\end{tabular}
	}}
	\vspace{-0.2cm}
	\caption{From top left to bottom right: Floorplan of the real world environment. The HSR robot in the office environment. Example episodes in the real world environment, the (found) target objects are shown in grey. Bottom right: failure case, the agent moves repeatedly back and forth between two rooms until it reaches the timeout.}
  	\label{fig:real_world}
\end{figure}
\setlength{\tabcolsep}{6pt}
\renewcommand{\arraystretch}{1}

\subsection{Real World Experiments}
To test the transfer to the real world, we transfer the policy onto a Toyota HSR robot. While the HSR has an omnidirectional drive, we restrict its motion to match that of the LoCoBot's differential drive in simulation. We run the experiments in our office building, representing a common office environment. We use rooms covering roughly a size of 320~square~meters, excluding small corners that cannot be navigated safely. We assume to have access to an accurate semantic camera. For this, we use the robot's depth camera to construct a map of the explored environment and overlay it with a previously semantically annotated map. In each episode, we randomly spawn 1-6 virtual target objects in this map by adding them to the semantic overlay. The actions are computed on the onboard CPU and executed at roughly \SI{7}{\hertz}. We define a maximum episode length of 6 minutes, roughly matching simulation. The robot is equipped with bumper sensors in its base that stop it upon any collision, in which case we deem the episode unsuccessful. \figref{fig:real_world} shows the real-world setup.

We make the following adaptations for the real world:
To minimize collisions, we inflate the map by \SI{5}{\centi\meter} and scale the actions of the agent by a factor of 0.55. Secondly, we reduce the temperature of the softmax activation of the long-horizon predictions to 0.1. We find that this increases the agent's confidence in its own predictions and leads to more target-driven exploration. We evaluate the agent for a total of 40 episodes, spread across different numbers of target objects. The results from this experiment are presented in \tabref{tab:real_world}. We observe that the agent successfully transfers to the real world, bridging differences in the robot's motion model, sensors, and environment layouts. Overall it solves 55\% of all episodes successfully with almost no decrease as the number of target objects increases. We find two main difficulties in the real world: while generally navigating smoothly, the agent occasionally collides with door frames or small objects such as posts. This may be caused by the mismatch in the robot's motions and controllers as well as due to the training environments consisting of mostly clean edges and little unstructured clutter. Secondly, in a very small number of episodes, the agent gets stuck moving back and forth between close spots, as the long-horizon prediction keeps changing back and forth. These results suggest a potential to further increase the success in the real world by finetuning the learned agent on the real robot, in particular to further reduce the number of collisions. Example trajectories from this experiment are shown in \figref{fig:real_world} and in the accompanying video.

\section{Conclusion}\label{sec:conclusion}
In this paper, we proposed a novel reinforcement learning approach for object search that unifies short- and long-horizon reasoning into a single model. To this end, we introduced a multi-object search task in continuous action space and formulated an explicit prediction task which allows the agent to guide itself over long-horizons. We demonstrated that our approach significantly outperforms other learning-based methods which struggle to perform efficient long-term exploration in this continuous space. By focusing on structured semantic map inputs, our approach learns complex exploration behaviors in comparably few steps and generalizes effectively to unseen apartments. Moreover, we successfully transferred the approach to the real world and find that the agent bridges the sim-real gap and exhibits the potential for further improvement if given the chance to adapt to the motion model of the real robot.

In the future, we aim to further exploit the ability of the approach to learn different actions at a high control frequency. Particular, we will investigate the ability to incorporate control of the head camera which should further improve the agent's success rates. Furthermore, we are interested in the application to mobile manipulation in which the agent has to simultaneously navigate and control its arms~\cite{honerkamp2021learning}. A third promising direction is the ability of learning-based approaches to incorporate data-driven knowledge such as correlations between semantic classes in real-world environments. Training on much larger environments will provide exciting avenues to exploit this.

\footnotesize
\bibliographystyle{spmpsci}
\bibliography{biblio}








\end{document}